\documentclass{article}

\usepackage{microtype}
\usepackage{graphicx}
\usepackage{subcaption}
\usepackage{booktabs} %
\usepackage{amsthm}
\usepackage{url}
\usepackage{enumitem}
\usepackage{bbm}
\usepackage{lipsum}
\usepackage{graphicx}
\usepackage[most]{tcolorbox}
\usepackage{soul}
\usepackage{caption}
\usepackage{subcaption}
\usepackage{wrapfig}
\usepackage{lipsum}
\usepackage{enumitem}
\usepackage{multirow}
\usepackage{tikz}
\usepackage{scalerel}
\usepackage{mathtools}
\usepackage{booktabs}
\usepackage{amsmath}
\usepackage{xcolor}
\usetikzlibrary{arrows.meta}
\usepackage{basic_commands}
\usepackage{amsmath}
\usepackage{xcolor}
\usetikzlibrary{arrows.meta}
\usepackage{basic_commands}
\usepackage{tabularew}
\usepackage[endLComment=,italicComments=false]{algpseudocodex}

\AddToHook{cmd/ttfamily/after}{\frenchspacing}

\usepackage{hyperref}

\usepackage[preprint]{icml2026}

\usepackage{amsmath}
\usepackage{amssymb}
\usepackage{mathtools}
\usepackage{amsthm}
\usepackage{basic_commands}

\usepackage[capitalize,noabbrev]{cleveref}

\theoremstyle{plain}

\theoremstyle{definition}

\theoremstyle{remark}

\definecolor{myred}{HTML}{F54254}
\definecolor{myorange}{HTML}{FFB135}
\definecolor{mygreen}{HTML}{10BD35}
\definecolor{myblue}{HTML}{598BE7}
\definecolor{mypurple}{HTML}{907be3}
\definecolor{plgray}{HTML}{999999}
\renewcommand{\tt}[1]{\texttt{#1}}
\renewcommand{\phi}{\varphi}

\hypersetup{urlcolor=mypurple,citecolor=mypurple,linkcolor=mypurple}

\newcommand{\methodname}{RQL\xspace}

\def\wt{\widetilde}

\icmltitlerunning{Reversal Q-Learning}

\begin{document}

\twocolumn[
  \icmltitle{Reversal Q-Learning}

  \icmlsetsymbol{equal}{*}

  \begin{icmlauthorlist}
    \icmlauthor{Aditya Oberai}{sch}
    \icmlauthor{Seohong Park} {sch}
    \icmlauthor{Sergey Levine}{sch}
  \end{icmlauthorlist}

  \icmlaffiliation{sch}{University of California, Berkeley}

  \icmlcorrespondingauthor{Aditya Oberai}{aoberai@berkeley.edu}

  \icmlkeywords{Machine Learning, ICML}

  \vskip 0.3in
]

\printAffiliationsAndNotice{}  %

\begin{abstract}
Iterative generative modeling techniques, such as flow matching,
provide powerful tools to model complex behaviors for effective offline reinforcement learning (RL).
In this work, we propose a new off-policy RL algorithm that trains a flow policy based on prior data.
Our idea starts from the ``expanded'' Markov decision process (MDP) framework,
which treats individual flow refinement steps as separate actions in an MDP.
To enable off-policy RL within this framework, we apply two techniques:
we generate virtual on-policy trajectories (by ``reversing'' flows) to make this framework compatible with prior data,
and we apply a bias-and-variance reduction technique to mitigate the curse of horizon in off-policy RL.
We call the resulting algorithm \textbf{reversal Q-learning (RQL)}.
RQL has several advantages over previous flow-based RL methods:
it does not suffer from backpropagation through time, makes better use of the learned value function, and directly trains the full, expressive flow policy.
Through our experiments on $50$ challenging simulated robotic tasks,
we show that RQL leads to the best average offline RL performance compared to state-of-the-art flow-based offline RL algorithms.

\vskip 0.1in

\textbf{Code}: \url{https://github.com/aoberai/rql}

\textbf{Website}: \url{https://aober.ai/rql}

\end{abstract}

\begin{figure}[!t]
    \centering
    \includegraphics[width=\columnwidth]{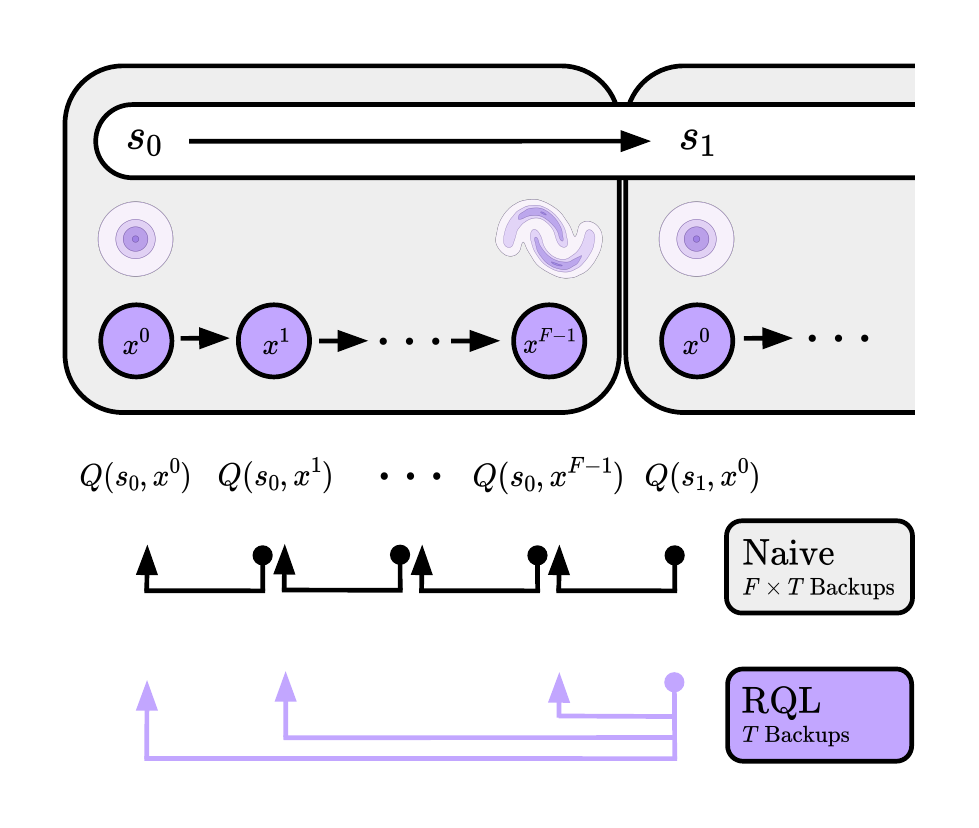}
    \caption{
    \footnotesize
      \textbf{Reducing the effective horizon.} 
        We reduce the effective TD horizon to leverage the expanded MDP framework for \emph{off-policy} RL. We avoid the naive solution which requires  $F \times T$ backups. Instead, RQL on average requires just $T$. 
    }
    \label{fig:tf2t_backups}
\end{figure} 
\section{Introduction}

Recent advancements in iterative generative modeling,
such as denoising diffusion~\citep{diffusion_sohl2015, ddpm_ho2020} and flow matching~\citep{flow_lipman2023, flow_albergo2023, flow_liu2023},
have provided powerful tools for effective off-policy
reinforcement learning (RL)~\citep{dql_wang2023,idql_hansenestruch2023, fql_park2025}.
By modeling complex behaviors in offline datasets with expressive generative models (e.g., by training diffusion or flow policies),
they can capture diverse behavioral priors that can be rapidly adapted to downstream tasks.

While promising in principle, training diffusion or flow policies with off-policy RL is a difficult problem.
This challenge stems from their \emph{iterative} nature.
For example, if we na\"ively train a diffusion policy to maximize a learned value function~\citep{ddpg_lillicrap2016, dql_wang2023},
gradients are backpropagated through the entire iterative generative process,
often leading to unstable training and suboptimal performance~\citep{fql_park2025}.
Prior work has sidestepped this issue using other techniques
like weighted regression~\citep{qipo_zhang2025}, distillation~\citep{fql_park2025}, and rejection sampling~\citep{idql_hansenestruch2023}, but these approaches come with their own limitations (see \Cref{sec:related}).

In this work, we consider an alternative paradigm that has recently been explored in diffusion-based \emph{on-policy} RL~\citep{ddpo_black2023, dpok_fan2023, dppo_ren2025}.
The idea is simple: instead of treating a diffusion policy as a black box that generates an action from a state,
this paradigm treats individual denoising steps as part of a Markov decision process (MDP),
effectively expanding the horizon by $F$ times (\Cref{fig:tf2t_backups}).
This way, we can fully avoid handling tricky issues from training iterative policies with RL,
such as backpropagation through time.
Previous work has shown that this expanded MDP paradigm is highly effective
when combined with on-policy algorithms like REINFORCE~\citep{reinforce_williams1992} and PPO~\citep{ppo_schulman2017}.

Unfortunately, this expanded MDP framework is not directly suitable for \emph{off-policy} RL,
whose goal is to train diffusion or flow policies with RL in a sample-efficient manner, leveraging prior data.
There are two main reasons.
First, standard offline datasets only contain state-action pairs from the original environment,
and do not provide diffusion or flow trajectories corresponding to the expanded MDP.
Second, the MDP expansion increases the horizon by $F$ times,
which makes it challenging to estimate accurate values due to ``the curse of horizon'' in off-policy RL~\citep{horizon_liu2018, sharsa_park2025}.

Our key insight in this work is that the \emph{reversibility} of \emph{deterministic} iterative generative models (e.g., flow matching)
provides an effective solution to both challenges.
Specifically, we first generate ``virtual'' trajectories in the expanded MDP,
by reconstructing the flow trajectories that the current policy would have produced for each state-action pair in the dataset.
This is done by solving an inverse problem via reverse flows.
We then apply multi-step returns to these virtual trajectories to reduce the effective horizon for value function learning.
Since these virtual trajectories are fully deterministic and on-policy,
we can obtain unbiased and zero-variance return estimates from otherwise biased multi-step returns~\citep{rl_sutton2005}.

We call the resulting off-policy flow RL algorithm \textbf{reversal Q-learning (RQL)},
which is the main contribution of this work.
Through our diverse experiments across $50$ simulated robotic tasks,
we demonstrate that RQL leads to the best performance compared to a number of strong off-policy flow-based RL baselines.
We show that RQL is particularly strong in challenging long-horizon manipulation and locomotion environments.

\section{Related Work}
\label{sec:related}

\textbf{RL with iterative generative models.}
Prior works have developed a variety of techniques to use modern iterative generative models (e.g., denoising diffusion~\citep{diffusion_sohl2015, ddpm_ho2020} and flow matching~\citep{flow_lipman2023, flow_albergo2023, flow_liu2023})
for data-driven RL, such as offline RL~\citep{batch_lange2012, offline_levine2020} and offline-to-online RL.
These works have employed diffusion or flow matching for trajectory modeling~\citep{diffuser_janner2022, dd_ajay2023, guidedflow_zheng2023, hdmi_li2023, hd_chen2024},
world modeling~\citep{synther_lu2023, dwm_ding2024, pgd_jackson2024, diamond_alonso2024},
and policy learning~\citep{dql_wang2023, idql_hansenestruch2023, sfbc_chen2023, edp_kang2023, dppo_ren2025, fql_park2025}.
Our work falls into the last category.
We aim to develop a better algorithm to train a flow policy for off-policy RL leveraging prior data~\citep{rlpd_ball2023}.

\textbf{RL with diffusion and flow policies.}
Due to their iterative nature, training diffusion or flow policies with RL is not a straightforward task.
A number of diverse approaches have been proposed to guide the iterative generation process to maximize returns.
These methods are based on different principles,
such as backpropagation through time~\citep{dql_wang2023, diffcps_he2023, consistencyac_ding2024, srdp_ada2024, entropydql_zhang2024, sorl_espinosa2025},
regression~\citep{qgpo_lu2023, edp_kang2023, idql_hansenestruch2023, sfbc_chen2023, qvpo_ding2024, qipo_zhang2025},
distillation~\citep{fql_park2025, floq_agrawalla2025},
MDP expansion~\citep{dppo_ren2025, bdpo_gao2025},
and more~\citep{dipo_yang2023, parl_mark2024, dac_fang2025, dsrl_wagenmaker2025, sacflow_zhang2025}.
In the rest of this section,
we discuss why some of these paradigms may be limited in practice and how our method can provide a better alternative.

\textbf{(1) Backpropagation through time.}
Arguably, the most straightforward way to train a diffusion policy with RL is
to directly maximize a learned value function with gradient ascent, treating the iterative generation process as a black box.
While prior work has shown that this can sometimes be effective~\citep{dql_wang2023, diffcps_he2023, srdp_ada2024, entropydql_zhang2024},
this paradigm often suffers from an issue called backpropagation through time (BPTT), especially when using larger iteration steps.
Since gradients are propagated through the long chain of the entire iterative generation procedure,
it often causes training instability and leads to suboptimal performance in practice~\citep{fql_park2025}.
In contrast, our method does not suffer from the BPTT issue
because we treat iterative refinement steps as distinct MDP environment steps.

\textbf{(2) Regression.}
To avoid BPTT, many previous works have explored regression-based techniques to maximize returns with diffusion or flow policies.
These methods include weighted regression~\citep{qgpo_lu2023, edp_kang2023, qvpo_ding2024, qipo_zhang2025},
rejection sampling~\citep{sfbc_chen2023, idql_hansenestruch2023, aligniql_he2024, sharsa_park2025}, and filtering~\citep{cfgrl_frans2025, pi06_intelligence2025}.
While these approaches do not suffer from the instabilities of BPTT,
they only use zeroth-order information from the value function
(i.e., they do not use value gradients),
which often leads to suboptimal performance (in weighted regression- or filtering-based methods)
or requires a large amount of compute (in rejection sampling-based methods)~\citep{bottleneck_park2024}.
Unlike these regression-based approaches,
we make better use of the value function by utilizing its first-order (gradient) information,
which we show leads to better performance in practice.

\textbf{(3) MDP expansion.}
An alternative paradigm to diffusion policy learning is to treat iterative refinement steps as MDP steps,
and solve this ``expanded'' MDP with a standard, off-the-shelf RL algorithm.
This framework is beneficial in that it does not suffer from BPTT and can fully utilize value gradients.
Prior work has shown that variants of this idea indeed lead to strong performance in on-policy RL settings~\citep{ddpo_black2023, dpok_fan2023, dppo_ren2025}.
However, this framework has rarely been applied to an offline \emph{off-policy} RL setting.
This is mainly because
(1) the original dataset does not contain full diffusion trajectories and
(2) it makes the horizon $F$ times longer (where $F$ is the number of iterative refinement steps),
which exacerbates ``the curse of horizon'' in off-policy value learning~\citep{horizon_liu2018, sharsa_park2025}.

To our knowledge, the only prior work that applies MDP expansion to off-policy RL is BDPO~\citep{bdpo_gao2025},
which concerns stochastic diffusion policies and employs bi-level hierarchical value functions to deal with the increased horizon.
Unlike BDPO, our method is based on deterministic ``reverse'' flows,
which enables us to address the horizon challenge without potentially complicated hierarchies.
Empirically, we also show that RQL leads to substantially better performance than this prior work.

\section{Preliminaries}
\label{sec:prelim}

\textbf{Problem setting.}
We consider a Markov decision process (MDP) defined as $\gM = (\gS, \gA, r, \mu, p)$~\citep{rl_sutton2005}.
$\gS$ is a state space, $\gA$ is an action space, $r(s, a): \gS \times \gA \to \sR$ is a reward function,
$\mu(s) \in \Delta(\gS)$ is an initial state distribution, and $p(s' \mid s, a): \gS \times \gA \to \Delta(\gS)$ is a transition dynamics kernel,
where $\Delta(\gX)$ denotes the set of probability distributions over a space $\gX$.
We also assume that we are given a prior dataset $\gD = \{\tau^{(n)}\}_{n \in \{1, 2, \ldots, N\}}$
consisting of trajectories $\tau = (s_0, a_0, r_0, s_1, \ldots, s_T)$,
which may correspond to human demonstrations, previous rollouts, or even suboptimal data.

In this work, we consider the problem of \emph{offline RL}.
That is, we aim to find a return-maximizing policy leveraging a prior dataset $\gD$.
Formally, our goal is to train a policy $\pi(a \mid s): \gS \to \Delta(\gA)$ (based on $\gD$)
that maximizes the discounted sum of rewards:
\begin{align}
J(\pi) = \E_{\tau \sim p^\pi(\tau)}\left[ \sum_{t=0}^\infty \gamma^t r(s_t, a_t) \right],
\end{align}
where $\gamma \in (0, 1)$ is a discount factor and
\begin{align}
p^\pi(\tau) = &\mu(s_0)\pi(a_0 \mid s_0)p(s_1 \mid s_0, a_0) \nonumber \\ &\cdots \pi(a_{T-1} \mid s_{T-1})p(s_T \mid s_{T-1}, a_{T-1}).
\end{align}

\textbf{Flow policies.}
Flow matching~\citep{flow_lipman2023, flow_albergo2023, flow_liu2023} provides a scalable way
to train an expressive generative network to model a continuous data distribution.
In this work, we consider \emph{flow policies},
which model action distributions via flow matching.
Formally, a flow policy is modeled by
a time-dependent velocity field $v(s, x, f): \gS \times \sR^d \times [0, F] \to \sR^d$,
where we assume $\gA = \sR^d$ and use $f$ to denote the time variable
(this choice is to avoid a notational clash with the time step $t$ in the MDP),
and $F$ denotes the maximum time.
This velocity field induces a \emph{flow} $\psi(s, x, f): \gS \times \sR^d \times [0, F] \to \sR^d$,
which is defined as the unique solution~\citep{smooth_lee2002}
to the following ordinary differential equation (ODE):
\begin{align}
\frac{\de}{\de f} \psi(s, x, f) = v(s, \psi(s, x, f), f).
\end{align}
This ODE transforms a fixed prior distribution (e.g., the standard normal $\gN(0, I_d)$) at $f = 0$
into a different distribution at $f = F$, which defines the action distribution of the flow policy.
In practice, we use the Euler method with $F$ iteration steps to solve the ODE for $\psi$:
i.e., we compute
\begin{align}
x^{f+1} \gets x^f + v(s, x^f, f)
\end{align}
at $f = 0, 1, \ldots, F-1$, where $x^0$ is sampled from the prior distribution
and $x^F$ approximates the output $\psi(s, x^0, F)$.

Flow policies are often trained to model behavioral action distributions in the dataset $\gD$.
This can be done by minimizing the following \emph{flow-matching} loss:
\begin{align}
\gL^\mathrm{BC}(v) = \E_{\substack{(s, a) \sim \gD, \\ x^0 \sim \gN(0, I_d), \\ f \sim \gU(0, F)}}
\left[ \| v(s, x^f, f) - \frac{1}{F}(a - x^0) \|_2^2 \right], \label{eq:bc_flow}
\end{align}
where
\begin{align}
x^f = (1 - f/F)x^0 + (f/F)a
\end{align}
and $\gU$ denotes a uniform distribution.
It has been shown that the resulting velocity field generates a flow
that transforms the Gaussian distribution $\gN(0, I_d)$
into the behavioral action distribution $\pi^\beta(a \mid s)$ of the dataset~\citep{flow_lipman2024, pi0_black2024, fql_park2025}.

\section{Reversal Q-Learning}

Our goal in this work is to develop a performant (offline) off-policy RL algorithm that leverages prior data to train a flow policy.
As briefly discussed in \Cref{sec:related},
we develop our method based on the expanded MDP framework.

In this section, we first formally define the expanded MDP framework in \Cref{sec:expanded}
and describe why it is challenging to directly apply this framework to off-policy RL in \Cref{sec:challenges}.
Then, we introduce our method, \textbf{reversal Q-learning (RQL)}, as a solution in \Cref{sec:solution},
and discuss practical implementation techniques in \Cref{sec:impl}.

\label{sec:expanded}
\begin{figure}[h!]
    \centering
    \includegraphics[width=\columnwidth]{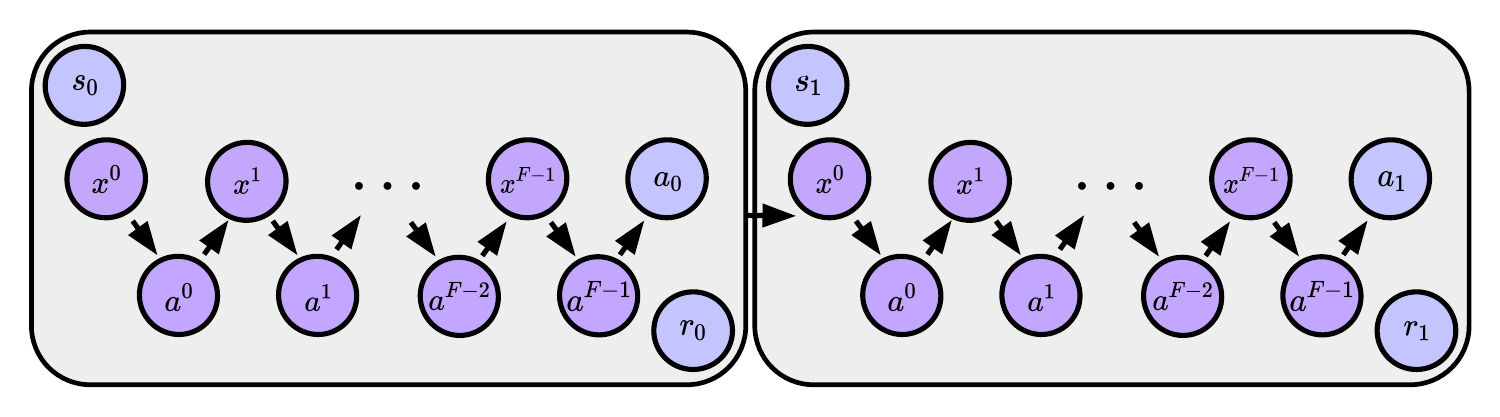}
    \caption{
    \footnotesize
    \textbf{Expanded MDP.}
    The expanded MDP construction treats individual denoising steps as individual actions, which enables training a diffusion or flow policy with a standard RL algorithm. F denotes the number of diffusion or flow integration steps.
    }
    \label{fig:expanded}
\end{figure}
\subsection{Expanded MDPs}

The main idea behind the expanded MDP framework
is to treat each Euler integration step in a flow policy as a separate action.
Essentially, this ``expands'' the MDP horizon by $F$ times,
where $F$ is the number of Euler integration steps.
This expanded MDP framework was originally proposed
in prior work in diffusion models and diffusion policies~\citep{ddpo_black2023, dpok_fan2023, dppo_ren2025}.
In this work, we consider flow policies instead of diffusion policies,
and we describe its (deterministic) flow variant in this section.

Specifically, given an MDP $\gM=(\gS, \gA=\sR^d, r, \mu, p)$,
the expanded MDP is defined as $\wt \gM = (\wt \gS, \gA=\sR^d, \wt r, \wt \mu, \wt p)$.
The augmented state space $\wt \gS = \gS \times \sR^d \times \{0, 1, \ldots, F-1\}$ consists of
elements corresponding to the tuples $(s, x, f)$ of a state $s \in \gS$,
a partially generated action $x \in \sR^d$, and the discretized flow time $f \in \{0, 1, \ldots, F-1\}$.

The transition dynamics kernel $\wt p((s', x', f') \mid (s, x, f), a): \wt \gS \times \sR^d \to \Delta(\wt \gS)$ is defined as follows.
If $f < F - 1$, $(s', x', f')$ is deterministically set to $(s, x+a, f+1)$.
Otherwise, $s'$ is sampled from $p(\cdot \mid s, x+a)$,
$x'$ is sampled from $\gN(0, I_d)$, and $f'$ is deterministically set to $0$.
Intuitively, the expanded MDP queries the original MDP every $F$ steps to update the environment state $s$,
and otherwise only updates the partially generated action $x$ following the Euler integration rule.
The flow step $f$ in the expanded MDP serves as a counter.
For the initial state distribution $\wt \mu(s, x, f)$,
$s$ is sampled from $\mu(\cdot)$, $x$ is sampled from $\gN(0, I_d)$,
and $f$ is deterministically set to $0$.

The reward function $\wt r((s, x, f), a): \wt \gS \times \sR^d \to \sR$ is defined as follows:
if $f < F - 1$, $\wt r((s, x, f), a) = 0$ and otherwise $\wt r((s, x, f), a) = r(s, x+a)$.
In other words, as in the new transition dynamics,
rewards in the expanded MDP are given only every $F$ steps by querying the original reward function.
Similarly, only rewards from the original MDP are discounted, defined by a modified discount factor $\wt \gamma$: if $f < F - 1$, $\wt \gamma = 1$ and otherwise $\wt \gamma = \gamma$.  

\subsection{Challenges}
\label{sec:challenges}

The expanded MDP framework defined in the previous section allows us to use an existing RL algorithm
to directly train individual refinement steps of a flow policy to maximize returns.
Indeed, prior work has shown that (a diffusion-based variant of) this expanded MDP framework
enables training performant diffusion policies
when combined with \emph{on-policy} RL methods like PPO~\citep{ppo_schulman2017, dppo_ren2025}.

However, this framework in its vanilla form is not directly suitable for \emph{off-policy} RL,
which is the main focus of this work.
There are two challenges.
First, the given offline dataset $\gD = \{(s_0, a_0, r_0, s_1, \ldots, s_T)^{(n)}\}$
only consists of transitions in the original MDP
and does not contain flow trajectories for the expanded MDP.
In other words, we do not have intermediate flow integration steps for each $(s, a, r, s')$ tuple in the dataset.

Second, and perhaps more importantly, this expanded MDP framework increases the horizon length by $F$ times.
Compared to on-policy RL methods,
which can tolerate long horizons relatively well thanks to on-policy value estimation techniques (e.g., GAE~\citep{gae_schulman2016}),
off-policy RL struggles more as the horizon grows~\citep{horizon_liu2018}.
This is mainly because off-policy RL typically relies on temporal difference (TD) learning to estimate off-policy values,
where biases in TD targets accumulate over the entire horizon and can substantially harm performance~\citep{sharsa_park2025}.

\subsection{Solution: Reversal}
\label{sec:solution}

Our key insight in this work is that the \emph{reversibility} and \emph{determinism} of flow ODEs
provide us with a solution that addresses both of the challenges.
Specifically, for each $(s, a, r, s')$ tuple in the original dataset,
we first generate ``virtual'' flow trajectories in the expanded MDP
by following the flow ODE in the \emph{reverse} direction.
Next, we apply multi-step returns to these virtual trajectories to reduce the effective horizon.
Importantly, since the virtual flow trajectories are deterministic and on-policy,
these multi-step returns are unbiased and zero-variance, unlike in the general case~\citep{rl_sutton2005}.
\begin{figure}[h!]
    \centering
    \includegraphics[width=\columnwidth]{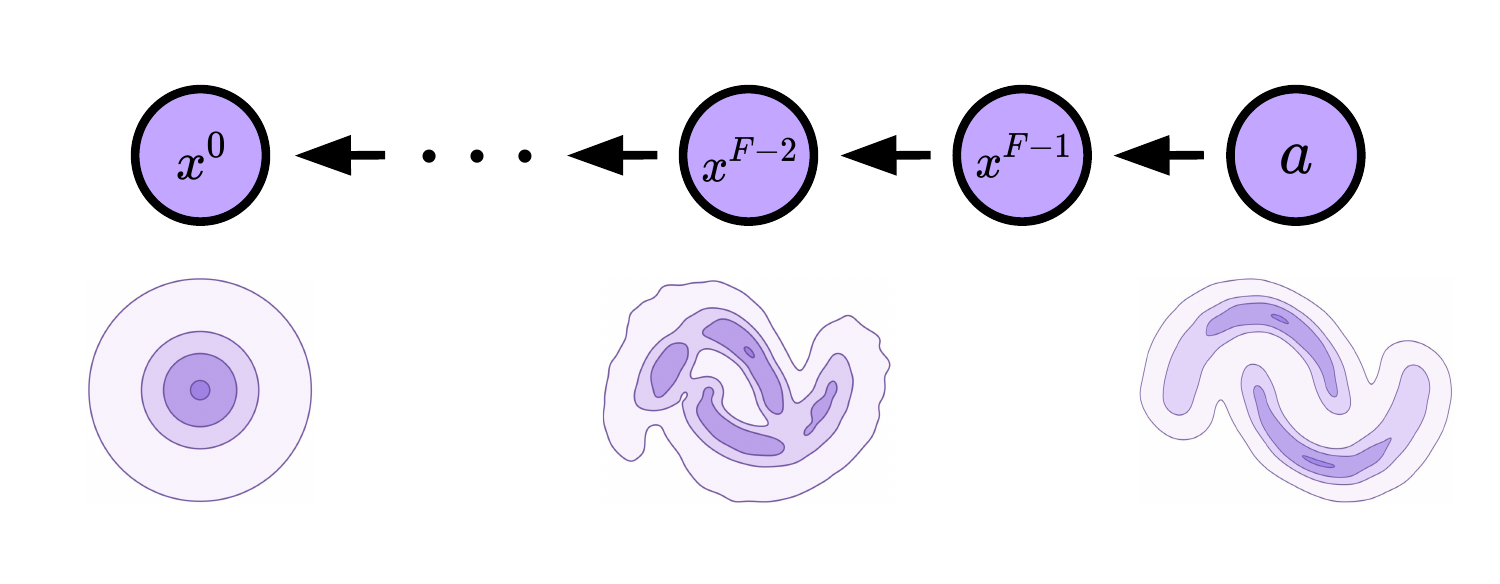}
    \caption{
    \footnotesize
    \textbf{Flow reversal.}
    We generate ``virtual'' on-policy flow trajectories
    by following the ODE in the reverse direction.
    }
    \label{fig:reversal}
\end{figure}

\textbf{Generating virtual trajectories for $\bm{\wt \gM}$.}
The first step is,
for a transition tuple $(s, a, r, s')$ in the dataset $\gD$,
to generate the corresponding flow trajectory in $\wt \gM$ with respect to the current flow policy $v$.
Our observation is that this can be done by computing the ``reverse'' flow $\theta(s, x, f): \gS \times \sR^d \times [0, F] \to \sR^d$
defined by the following ODE:
\begin{align}
\frac{\de}{\de f} \theta(s, x, f) = -v(s, \theta(s, x, f), f).
\end{align}
Note that the sign of the velocity field $v$ is reversed.
The reason behind this is that
a flow induces diffeomorphisms (i.e., smooth bijections whose inverses are also smooth)
between any two time steps under mild regularity assumptions~\citep{smooth_lee2002},
and its inverse flow is induced by the negative of the original velocity field.
This reverse ODE can be solved with the Euler method by computing
\begin{align}
x^{f-1} \gets x^f - v(s, x^f, f) \label{eq:reverse_ode}
\end{align}
at $f = F, F-1, \ldots, 1$, where the initial value is given by $x^F = a$ (\Cref{fig:tf2t_backups}).
After computing $x^0, x^1, \ldots, x^F$, we obtain the following transitions for $\wt \gM$:
\begin{align}
(\underbrace{(s, x^f, f)}_{\mathrm{state}}, \underbrace{x^{f+1} - x^f}_{\mathrm{action}}, \underbrace{0}_{\mathrm{reward}}, \underbrace{(s, x^{f+1}, f+1)}_{\mathrm{next\ state}}) \label{eq:expand_trans1}
\end{align}
for $f = 0, 1, \ldots, F-2$ and
\begin{align}
(\underbrace{(s, x^f, f)}_{\mathrm{state}}, \underbrace{x^{f+1} - x^f}_{\mathrm{action}}, \underbrace{r}_{\mathrm{reward}}, \underbrace{(s', x'^0, 0)}_{\mathrm{next\ state}}), \label{eq:expand_trans2}
\end{align}
for $f = F-1$, where $x'^0 \sim \gN(0, I_d)$.
We define $\wt \gD$ as the set of transitions in \Cref{eq:expand_trans1,eq:expand_trans2}.
Note that $\wt \gD$ depends on the current flow policy $v$,
and is re-computed from $\gD$ for each batch (see \Cref{alg:rql} for details).

\textbf{Reducing the effective horizon.}
With $\wt \gD$ defined above,
we can now train an off-policy value function for $\wt \gM$ with Q-learning.
For example, one may train a Q function $Q((s, x, f), a): \wt \gS \times \gA \to \sR$
with the following standard temporal difference loss:
\begin{align}
\gL(Q) = 
\E \Big[ \Big( &Q((s, x, f), a) - r \nonumber \\
&- \wt \gamma \max_{a'} \bar Q((s', x', f'), a')
\Big)^2 \Big],
\end{align}
where transition tuples $((s, x, f), a, r, (s', x', f'))$
are sampled from the dataset $\wt \gD$,
and $\bar Q$ denotes a target network~\citep{dqn_mnih2013}.
However, while this is a valid objective in theory,
it is challenging to obtain an accurate value function with this vanilla objective in practice.
This is because the horizon length has increased to $T \times F$ from $T$ in the expanded MDP,
where this increased horizon impedes off-policy value learning (see \Cref{sec:challenges} for a detailed explanation).

Our idea to address
this horizon challenge is
to observe that (some) multi-step returns are \emph{unbiased and zero-variance} in this expanded MDP,
because the flow trajectories in $\wt \gD$ are deterministic and on-policy.
Specifically, we propose to use the following multi-step Q-learning objective instead:
\begin{align}
\gL(Q) = \E_{\wt \tau} \Big[ \Big( &Q((s, x^f, f), a^f) -  r \nonumber \\
&- \gamma \max_{a'} \bar Q((s', x'^0, 0), a') \Big)^2 \Big], \label{eq:multistep}
\end{align}
where flow trajectories
\begin{align}
\wt \tau = \big(&\underbrace{(s, x^0, 0)}_{\mathrm{state}}, \underbrace{a^0}_{\mathrm{action}}, \underbrace{0}_{\mathrm{reward}},
\underbrace{(s, x^1, 1)}_{\mathrm{state}}, \underbrace{a^1}_{\mathrm{action}}, \underbrace{0}_{\mathrm{reward}}, \ldots, \nonumber \\
&\underbrace{(s, x^{F-1}, F-1)}_{\mathrm{state}}, \underbrace{a^{F-1}}_{\mathrm{action}}, \underbrace{r}_{\mathrm{reward}}, \underbrace{(s', x'^0, 0)}_{\mathrm{state}}\big)
\end{align}
are sampled from $\wt \gD$ and $f$ is sampled uniformly from $\{0, 1, \ldots, F-1\}$.

Intuitively, \Cref{eq:multistep} directly takes the TD target at the end of each flow trajectory,
skipping intermediate flow transitions (\Cref{fig:tf2t_backups}).
Since intermediate flow trajectories are fully deterministic and are synthesized with respect to the current policy $v$
(i.e., they are on-policy),
the multi-step return in \Cref{eq:multistep} is unbiased and zero-variance,
unlike standard off-policy multi-step TD learning~\citep{rl_sutton2005}.

This technique reduces the effective value horizon (i.e., the number of Bellman updates required to propagate information along each trajectory)
from $T \times F$ to $T$ on average.
In our experiments, we show that this ``value horizon reduction''~\citep{sharsa_park2025}
is indeed crucial in achieving strong performance in practice.

\subsection{Practical Algorithm}
\label{sec:impl}

Based on the ideas described in \Cref{sec:solution},
we now introduce a practical algorithm to train a flow policy with off-policy RL using prior data.
We call the resulting method \textbf{reversal Q-learning (RQL)}.

\textbf{Value learning.}
The multi-step value loss in \Cref{eq:multistep} requires computing the maximum over next actions ($\max_{a'}$).
Since na\"ively computing it often leads out-of-distribution queries of the Q-function,
especially when using offline data~\citep{offline_levine2020},
we instead use expectile regression~\citep{exp_newey1987, iql_kostrikov2022} to compute this maximum in an implicit manner.

Specifically,
we consider the following IVL-like loss (which is a value-only variant of implicit Q-learning~\citep{iql_kostrikov2022, ogbench_park2025})
to train a value function $V(s, x, f): \gS \times \sR^d \times \{0, \ldots, F-1\} \to \sR$:
\begin{align}
\gL(V) = \E_{\wt \tau}\big[
&\ell_2^\kappa\big(V(s, x^f, f) \nonumber \\
&- (r + \gamma V(s', x'^0, 0))\big)
\big], \label{eq:final_v}
\end{align}
where $\ell_2^\kappa(x) = |\kappa - \sI(x > 0)|x^2$ is the expectile loss with an expectile $\kappa$.
Intuitively, this asymmetric expectile loss approximates the $\max_{a'}$ operator in \Cref{eq:multistep}
without having to explicitly search for the maximum.

\textbf{Policy learning.}
To train a flow policy to maximize the learned value function $V$,
we employ a DDPG-style loss with a behavioral regularizer~\citep{ddpg_lillicrap2016, brac_wu2019, td3bc_fujimoto2021}.
Specifically, we train the velocity field $v$ to minimize the following loss:
\begin{align}
\gL(v) = &\underbrace{-\E_{\wt \tau}[V(s, x^f + v(s, x^f, f), f+1)]}_{\mathrm{value\ maximization}} \nonumber \\
&\underbrace{+\alpha\gL^\mathrm{BC}(v),}_{\mathrm{behavioral\ regularization}} \label{eq:final_pi}
\end{align}
where $\gL^\mathrm{BC}(v)$ is defined in \Cref{eq:bc_flow}
and $\alpha$ is a hyperparameter that controls the strength of the behavioral flow-matching regularizer.
Intuitively, the first term pushes velocity vectors to maximize the value,
while the second term regularizes the flow policy to be close to the prior dataset.

We found that having this behavioral regularizer (common in offline RL) is beneficial in practice for two reasons.
First, it allows the policy to capture useful behavioral priors from the dataset throughout the training.
Second, it encourages $x^0$ computed via reversal (\Cref{eq:reverse_ode})
to be closer to the prior distribution $\gN(0, I_d)$
by minimizing distributional shifts between the current policy and the prior dataset,
which in turn makes the target value in \Cref{eq:final_v} more accurate.

\textbf{Implementation.}
In practice,
we additionally apply action chunking~\citep{aloha_zhao2023, qc_li2025}
for both value functions and the flow policy,
which we found to improve performance.
We denote the resulting action-chunked dataset as $\gD_\mathrm{ac}$.
We summarize our algorithm in \Cref{alg:rql}.

\begin{algorithm}[t!]
\caption{Reversal Q-Learning (\methodname)}
\label{alg:rql}
\begin{algorithmic}

\State
\State Initialize value function $V(s, x^f, f)$, flow policy $v(s, x^f, f)$
\While{unfinished}
\State Sample $(s, a, r, s') \sim \gD_\mathrm{ac}$
\State Compute $\wt \tau$ using \Cref{eq:reverse_ode}
\State Train $V$ by minimizing $\gL(V)$ (\Cref{eq:final_v})
\State Train $v$ by minimizing $\gL(v)$ (\Cref{eq:final_pi})
\EndWhile

\vspace{5pt}

\end{algorithmic}
\end{algorithm}

\textbf{Why is RQL beneficial?}
The RQL algorithm described in \Cref{alg:rql} has several appealing properties.
First, by treating each individual flow step as a distinct action,
RQL does not suffer from backpropagation through time
or related issues that arise when performing RL with iterative models.
Second, the flow policy loss in \Cref{eq:final_pi} utilizes first-order (gradient) information from the value function,
which leads to better efficiency and performance compared to zeroth-order (value-gradient-free) methods,
such as the regression-based methods described in \Cref{sec:related}.
We empirically support this claim in our experiments.
Third, we reduce the increased effective horizon of the expanded MDP to the original length
\emph{without incurring any biases or variances}.
This mitigates the curse of horizon in off-policy RL,
enabling more accurate and effective off-policy value learning.

\section{Experiments}
\label{sec:exp}

In this section, we empirically evaluate the performance of RQL
on a variety of challenging simulated robotic manipulation tasks.
We also experimentally demonstrate
how each component of RQL is necessary to achieve strong performance in practice.

\begin{figure}[h!]
\centering
\hfill
\begin{subfigure}{0.19\linewidth}
    \centering
    \includegraphics[width=1.0\textwidth]{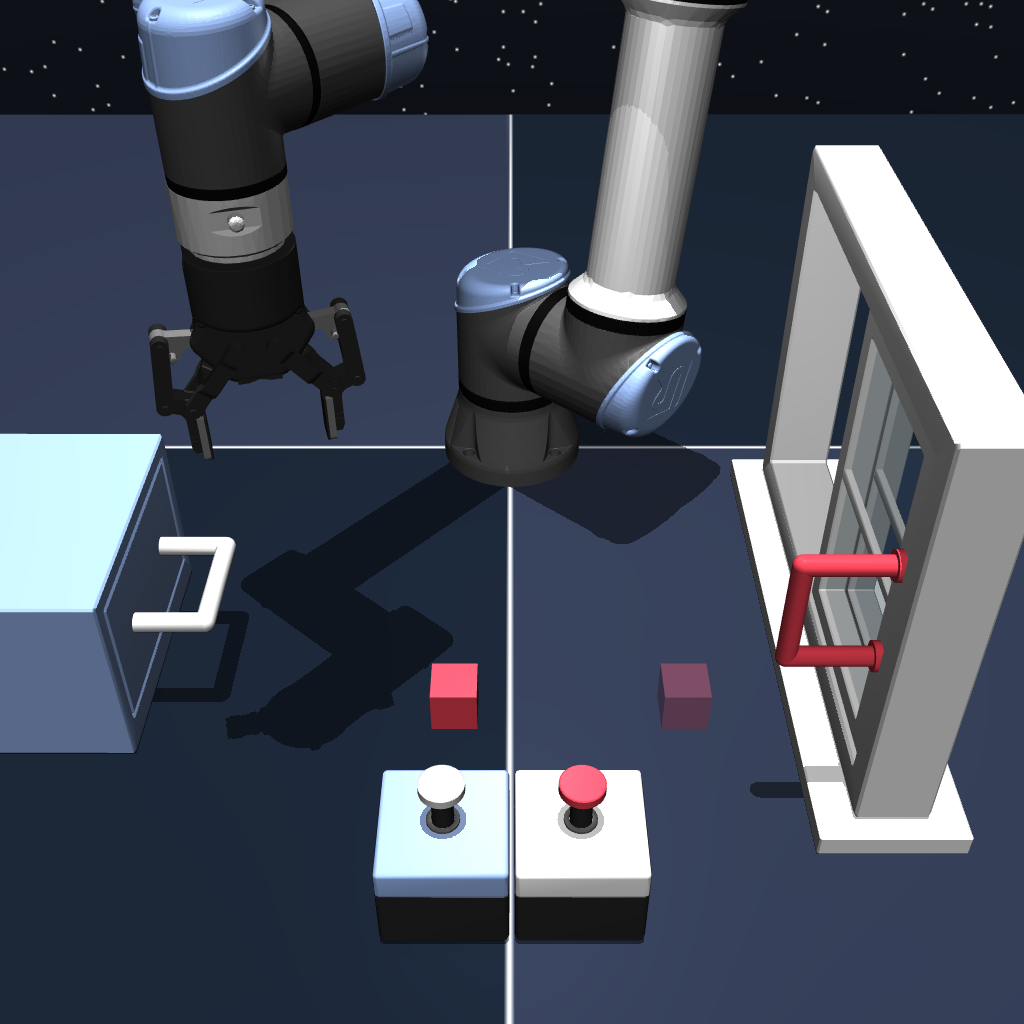}
    \caption*{\scriptsize\tt{scene}}
\end{subfigure}
\hfill
\begin{subfigure}{0.19\linewidth}
    \centering
    \includegraphics[width=1.0\textwidth]{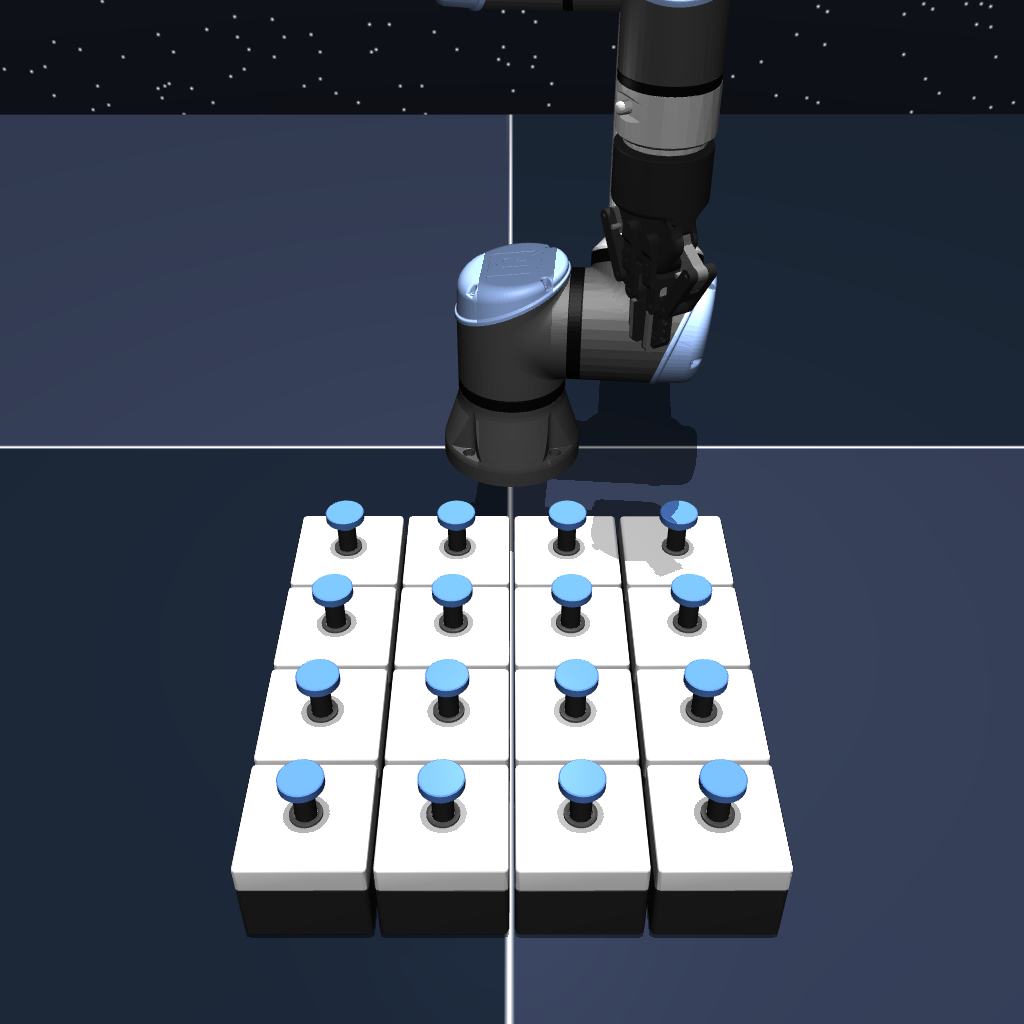}
    \caption*{\scriptsize\tt{puzzle-4x4}}
\end{subfigure}
\hfill
\begin{subfigure}{0.19\linewidth}
    \centering
    \includegraphics[width=1.0\textwidth]{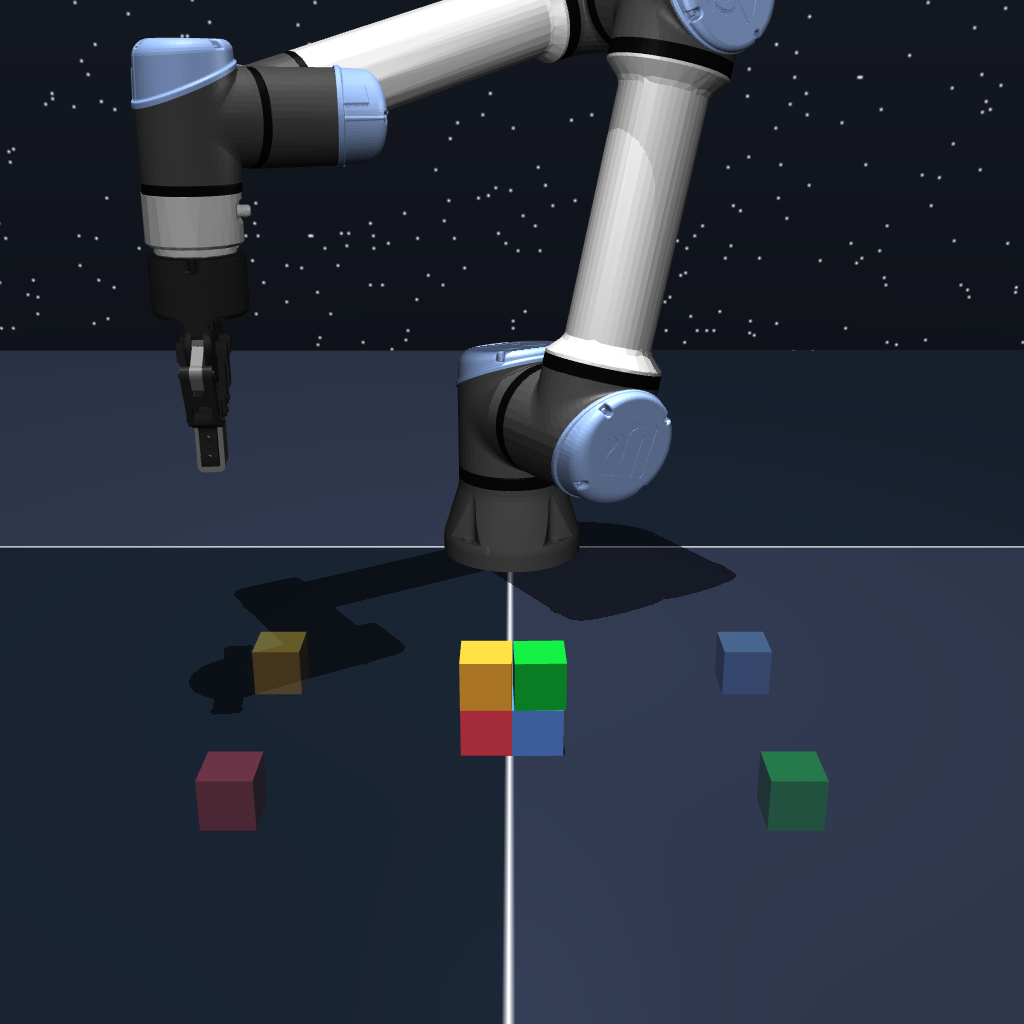}
    \caption*{\scriptsize\tt{cube-quad}}
\end{subfigure}
\hfill
\begin{subfigure}{0.19\linewidth}
    \centering
    \includegraphics[width=1.0\textwidth]{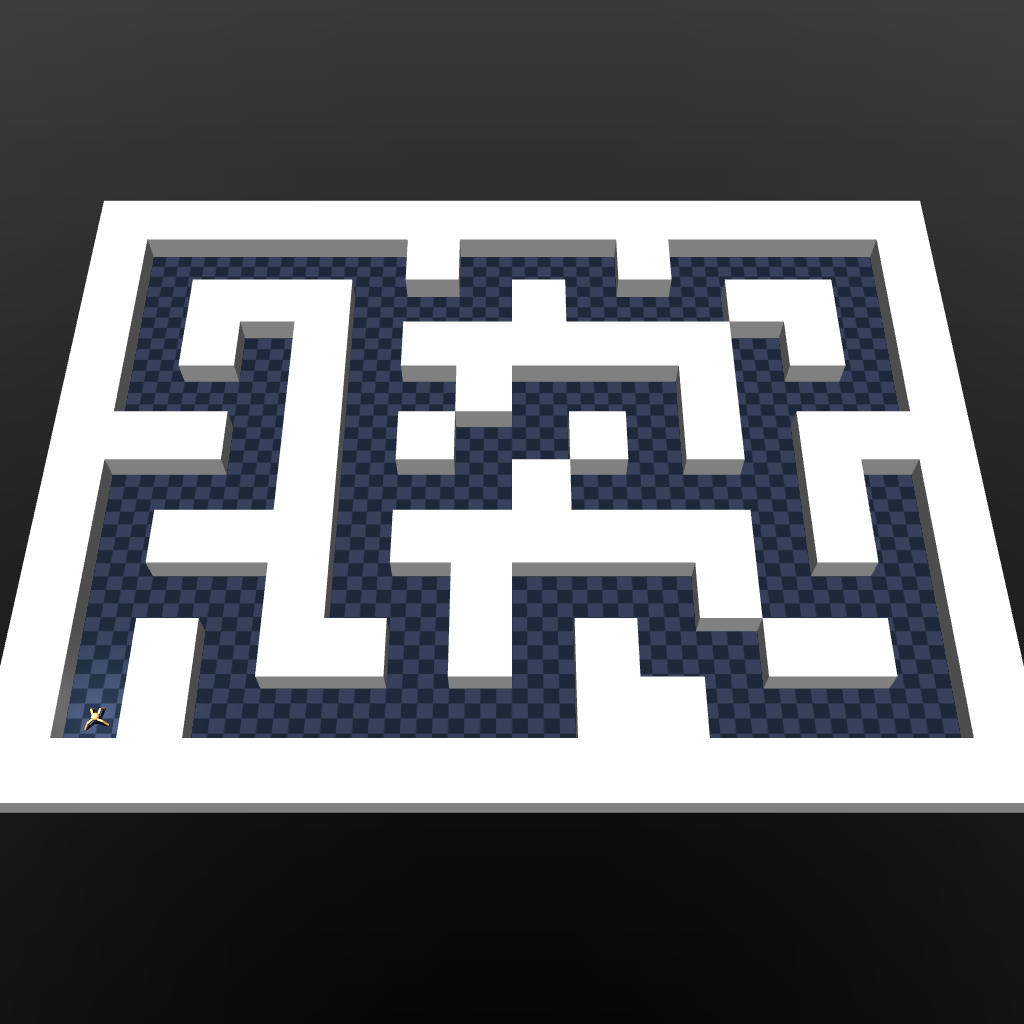}
    \caption*{\scriptsize\tt{amz-giant}}
\end{subfigure}
\hfill
\begin{subfigure}{0.19\linewidth}
    \centering
    \includegraphics[width=1.0\textwidth]{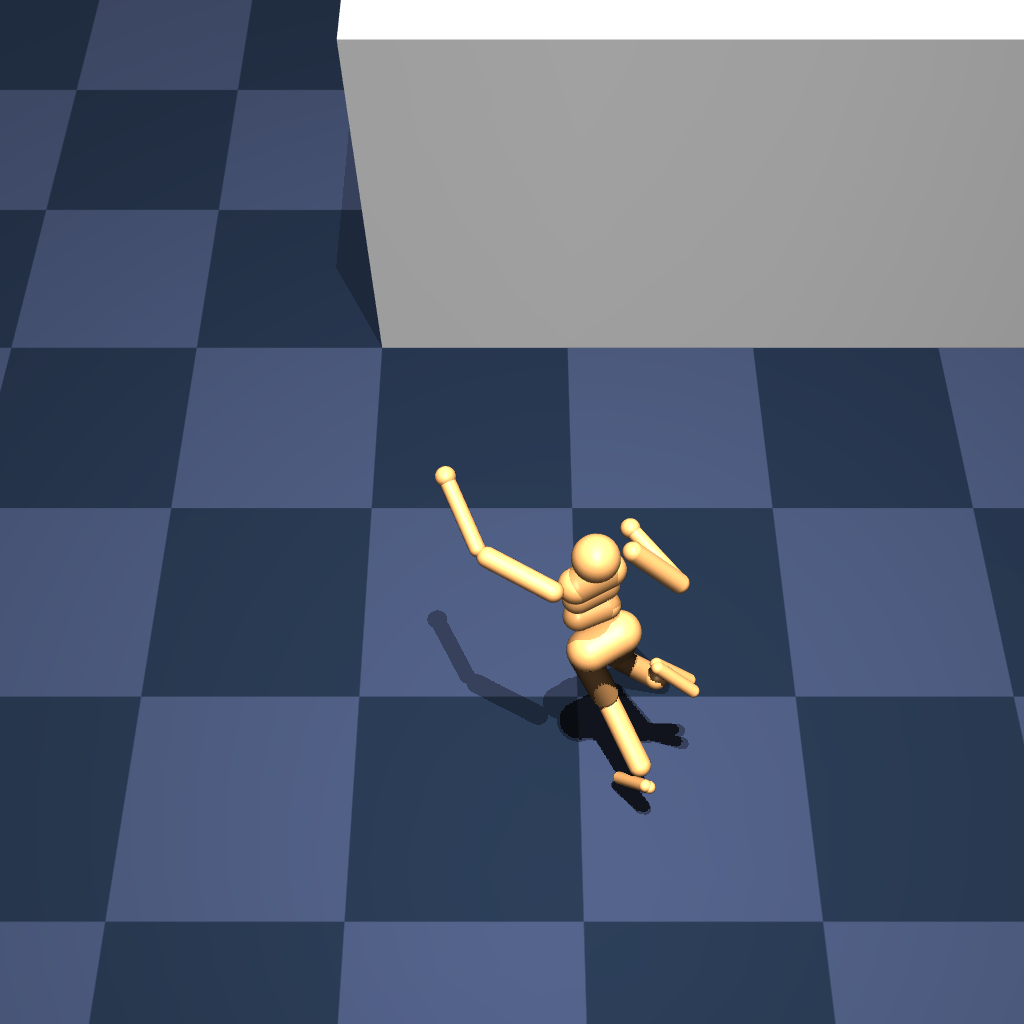}
    \caption*{\scriptsize\tt{hmz-large}}
\end{subfigure}
\vspace{5pt}
\caption{\textbf{Environments. }}
\label{fig:envs}
\end{figure}
\subsection{Experimental Setup}

\textbf{Tasks and datasets.}
We employ $50$ robotic manipulation tasks in the OGBench benchmark suite~\citep{ogbench_park2025} in our experiments.
Specifically, following \citet{qam_li2026}, we consider both manipulation tasks like \tt{scene}, \tt{puzzle}, and \tt{cube} as well as locomotion tasks like \tt{antmaze} and \tt{humanoidmaze} (\Cref{fig:envs}). 
These tasks require object manipulation with stitching (\tt{scene}), combinatorial reasoning (\tt{puzzle}), fine-grained control (\tt{cube}), long horizon locomotion (\tt{antmaze}), and high-dimensional control (\tt{humanoidmaze}). 
We consider two variants of tasks with different levels of difficulty
for \tt{puzzle}, \tt{cube}, \tt{antmaze}, and \tt{humanoidmaze} respectively.

Using the same experimental setting as ~\citet{qam_li2026}, we use the expanded 100M-sized datasets for \tt{cube-quadruple} and \tt{puzzle-4x4} provided by \citet{ogbench_park2025}, while other environments use the standard \tt{play} and \tt{navigate} datasets. These datasets consist of task-agnostic trajectories that repeatedly perform random atomic motions
(e.g., randomly press buttons in \tt{puzzle}). Hence, the agent must be able to stitch different trajectory segments in the dataset to solve the given task. In addition, we use the sparse reward variant of \tt{scene} and \tt{puzzle} following ~\citet{qam_li2026}, while other tasks use the standard semi-sparse reward function (i.e., one that only depends on the number of remaining tasks),
which is the default setting for OGBench \tt{singletask} tasks ~\citep{ogbench_park2025}. 

\textbf{Methods and comparisons.}
We compare RQL with diverse strong baselines across different categories.
For a Gaussian policy baseline, we consider ReBRAC~\citep{rebrac_tarasov2023}.
For flow-based off-policy RL baselines,
we consider $\mathbf{18}$ methods across seven categories:
FQL~\citep{fql_park2025} as a distillation-based method,
IFQL~\citep{idql_hansenestruch2023, fql_park2025} as a rejection sampling-based method,
FAWAC~\citep{awac_nair2020, fql_park2025} as a weighted regression-based method,
FBRAC ~\citep{fql_park2025} as a backpropagation through time method,
CGQL, DAC, and QSM ~\citep{qam_li2026, dac_fang2025, qsm_psenka2024} as test-time Q-gradient-based methods,
DSRL as a latent noise steering method ~\citep{dsrl_wagenmaker2025}, 
FEDIT as a residual edit method ~\citep{qam_li2026},
BAM and QAM ~\citep{qam_li2026} as adjoint matching methods, 
and BDPO ~\citep{bdpo_gao2025} as a method using the expanded MDP construction.

For methods other than BDPO and RQL,
we take the corresponding results from prior work~\citep{qam_li2026};
note that we use the same setting as this prior work, so the results are fully compatible.
We also use action-chunking variants of these methods to ensure a fair comparison.
All experiments in this work are averaged over four seeds, and we present $95\%$ confidence intervals in tables and figures.

\begin{figure}[h!]
    \centering
        \caption{
    \footnotesize
    \textbf{Overall Performance.}
    RQL exceeds the aggregate performance of all baselines across the 50 tasks.
    }
    \includegraphics[width=1.0\linewidth]{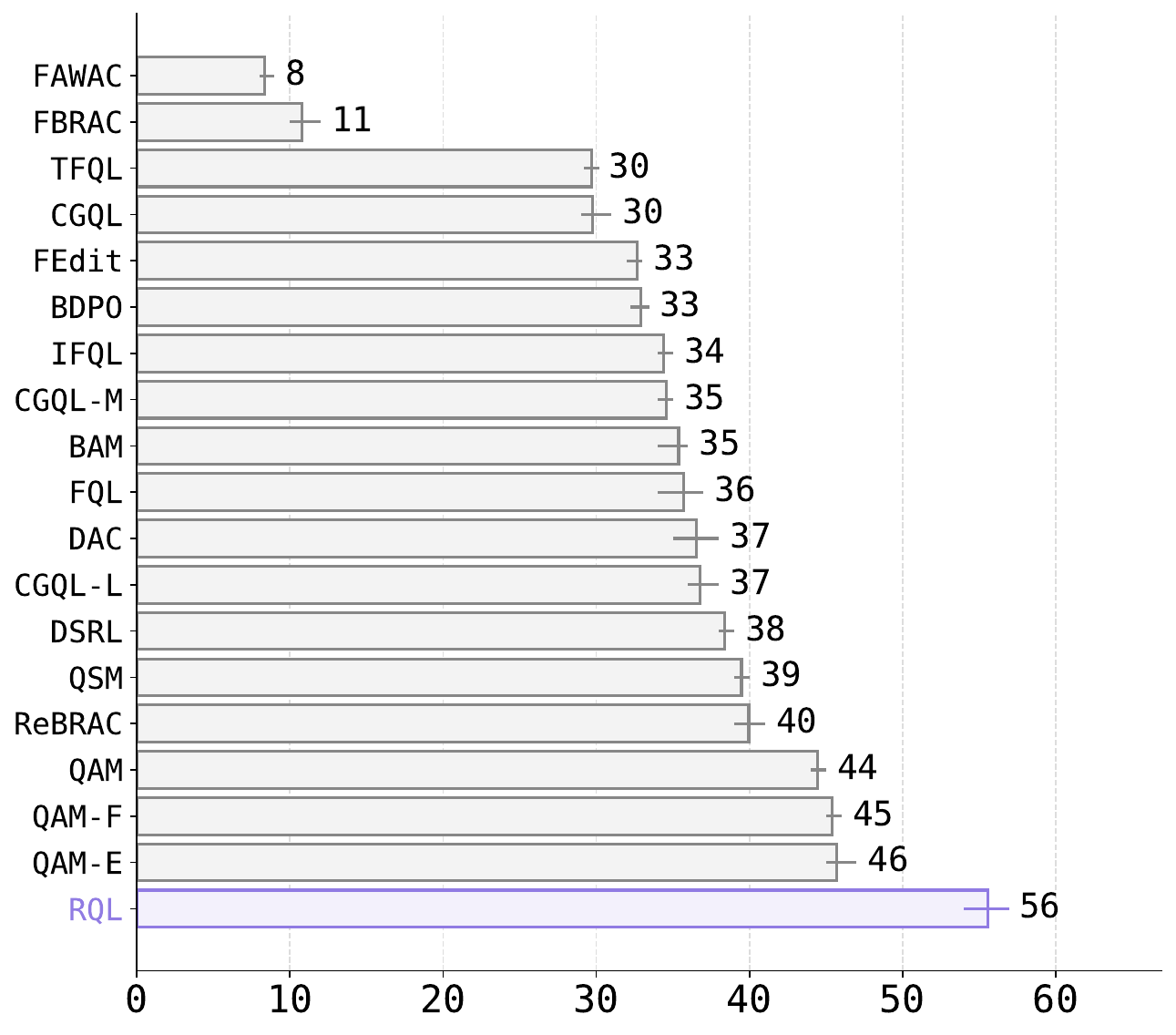}

    \label{fig:overall}
\end{figure}

\begin{figure*}[t!]
    \centering
    \includegraphics[width=0.9\linewidth]{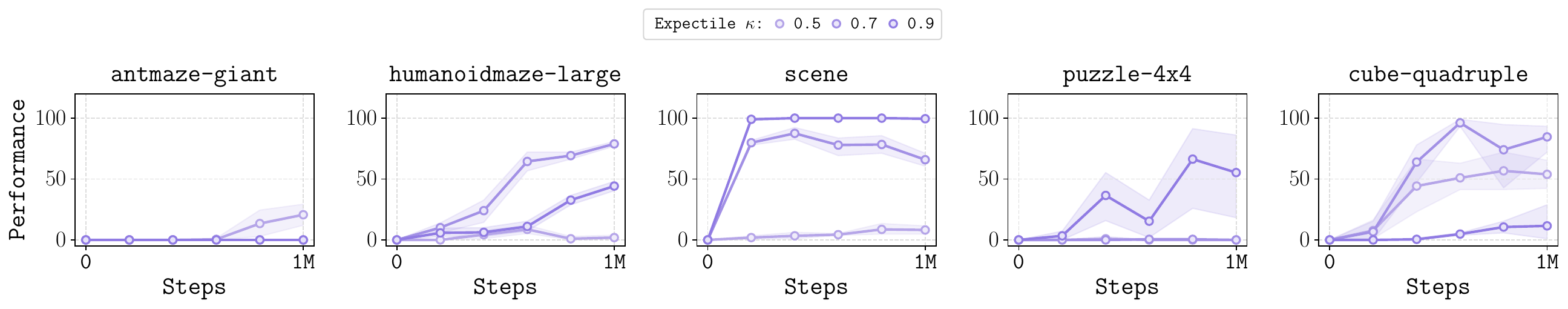}
    \caption{
    \footnotesize
      \textbf{Ablation study on the expectile $\kappa$.} 
    }
    \label{fig:expectile}
\end{figure*} 

\begin{figure*}[t!]
    \centering
    \includegraphics[width=0.9\linewidth]{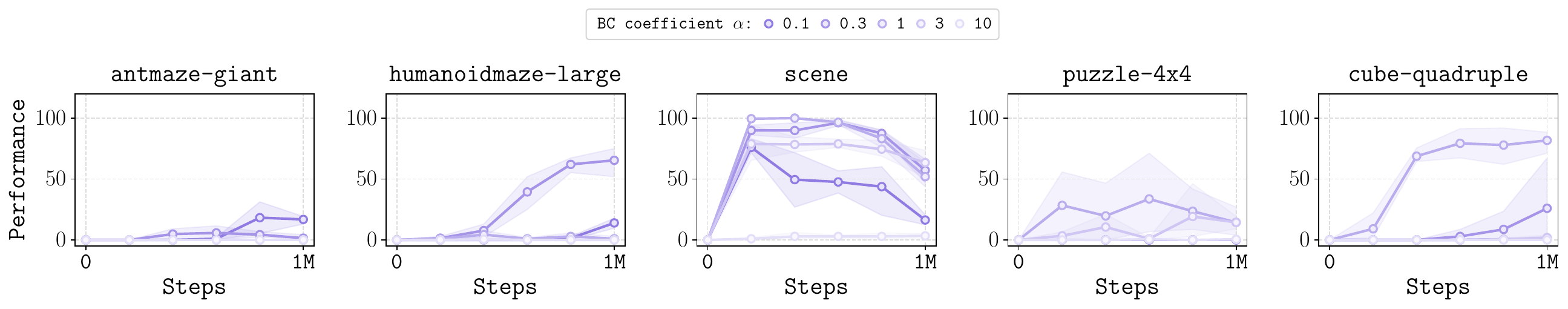}
    \caption{
    \footnotesize
      \textbf{Ablation study on the BC regularization coefficient $\alpha$.} 
    }
    \label{fig:alpha}
\end{figure*} 

\subsection{Results}

We present the full evaluation result on $50$ challenging robotic manipulation tasks in \Cref{table:main}
and aggregated performance in \Cref{fig:overall}.
The results show that RQL achieves the best average performance across the board,
showing particularly strong performance on the most challenging variants of tasks like \tt{antmaze-giant}, \tt{humanoidmaze-large}, \tt{puzzle-4x4}, and \tt{cube-quadruple}.
Note that our expanded MDP-based framework outperforms the most standard ways of training a flow policy from a value function,
such as backpropagation through time (FBRAC), rejection sampling (IFQL), and weighted regression (FAWAC).

One of our main ideas is to apply unbiased and zero-variance multi-step returns (\Cref{eq:multistep})
to reduce the effective horizon length for value learning. 
To understand the importance of value horizon reduction,
we compare the performance without this technique, denoted by TFQL in \Cref{fig:overall} and \Cref{table:main}. In particular, TFQL retains an effective horizon length of ``$T \times F$'', compared to ``$T$'' for RQL.
The results suggest that applying horizon reduction is indeed crucial in achieving strong performance,
and na\"ively using the vanilla expanded MDP framework (without horizon reduction)
can lead to a complete failure on some benchmark tasks.

\subsection{Ablation Study}

In this section, we ablate two components of RQL and discuss their effects. This is done on the singletask defaults of five representative tasks: \tt{antmaze-giant}, \tt{humanoidmaze-large}, \tt{scene}, \tt{puzzle-4x4}, and \tt{cube-quadruple}. 

\textbf{Expectile $\bm{\kappa}$.}
In RQL, we use the expectile loss (\Cref{eq:final_v}) to approximate the $\max$ operator in the Bellman operator.
\Cref{fig:expectile} shows an ablation study on the value of $\kappa$.
Note that $\kappa = 0.5$ corresponds to SARSA.
The results suggest that using a high $\kappa$ (i.e., ``more RL'') often leads to better performance across different tasks,
likely because the OGBench datasets are highly suboptimal for the given tasks.

\textbf{BC coefficient $\bm{\alpha}$.}
As in many other offline RL methods~\citep{td3bc_fujimoto2021, dql_wang2023, fql_park2025},
RQL also has a hyperparameter that interpolates between RL and BC (i.e., BC coefficient $\alpha$ in \Cref{eq:final_pi}).
We ablate this hyperparameter and present the results on tasks across diverse categories in \Cref{fig:alpha}.
The results show that $\alpha$ is the most important hyperparameter to tune.

\newpage

\section{Conclusion}

In this work, we proposed a flow-based off-policy RL algorithm based on the expanded MDP framework.
Our ideas based on ``flow reversal'' enable training an effective flow policy
without suffering from backpropagation through time or the curse of horizon in off-policy RL,
while making use of rich gradient information in the learned value function.
Through our experiments across a number of robotic manipulation tasks,
we empirically demonstrate that RQL achieves the best performance compared to other strong baselines,
especially on challenging long-horizon tasks.

\textbf{Limitations and future work.}
While RQL achieves strong empirical performance across diverse tasks,
it has several limitations, which open up diverse opportunities for future work.
First, we find offline RL performance is relatively sensitive to both the BC coefficient and expectile, and we expect that these hyperparameters need to be swept properly (\Cref{sec:tuning}) for best performance. 
Second, we only demonstrate the capabilities of RQL in the context of RL and robotic control.
Given the generality of this framework, we believe it may also be applied to fine-tune image generation models
or other modalities beyond RL and control, which we leave for future work. 

\newpage 
\section*{Acknowledgments}
This work was partly supported by the Korea Foundation for Advanced Studies (KFAS), AFOSR FA9550-22-1-0273, ONR N00014-25-1-2060, and DARPA ANSR. This research used the Savio computational cluster resource provided by the Berkeley Research Computing program at UC Berkeley.

\clearpage

\bibliography{example_paper}
\bibliographystyle{icml2026}

\newpage
\appendix
\onecolumn
\section{Full Result Table}
\label{sec: table}

\begin{table*}[h!]
    \caption{
\footnotesize
\textbf{Performance on $\bm{50}$ simulated robotic manipulation tasks.}
{\color{mypurple}{RQL}} generally achieves the best performance across the board,
particularly on more challenging, long-horizon tasks like \tt{humanoidmaze-large} and \tt{cube-quadruple}.
}
    \centering
    \makebox[\textwidth]{\scalebox{0.4}{
\begin{tabular}{clccccccccccccccccccc}
\toprule
 &  & \texttt{ReBRAC} & \texttt{FBRAC} & \texttt{BAM} & \texttt{FQL} & \texttt{FAWAC} & \texttt{CGQL} & \texttt{CGQL-M} & \texttt{CGQL-L} & \texttt{DAC} & \texttt{QSM} & \texttt{DSRL} & \texttt{FEdit} & \texttt{IFQL} & \texttt{QAM} & \texttt{QAM-F} & \texttt{QAM-E}  & \texttt{BDPO} & \texttt{TFQL} & \texttt{RQL} \\
\bottomrule
\end{tabular}}}
    \label{table:main}
\end{table*}

\newpage
\section{Additional Implementation Details}
\label{sec: details}

\textbf{Computing $\bm{x^f}$.}
For simplicity, \cref{sec:solution} shows that we compute $x^f$ at discretized flow intervals $f \in \{0, 1, \ldots, F-1\}$. In practice, we recognize that our value update rule in \cref{eq:final_v} does not require us to compute the entire flow trajectory but rather a singular $x^f$ for a flow time $f$. This $f$ can be sampled in many ways, we choose to partly sample $f$ from a continuous uniform distribution on the support $[0, F]$ and the other half uniformly within $\{0, 1, \ldots, F-1\}$. For these $f$, given that we have $x^F$, we want to compute $x^f$. This can be done with $F$ Euler steps as follows:
\begin{align}
x^{f-h} \gets x^f - h v(s, x^f, f) \label{eq:reverse_cf}
\end{align}
where $h$ is the fixed Euler step size $\frac{F-f}{F}$, iterated for $F$ steps from $x^F$, yielding $x^f$.

\textbf{Actor exponential moving average (EMA).}
We use an EMA of the flow policy during evaluation with $\lambda=0.999$ like previous works \citep{ddpm_ho2020, ema_karras2024}.
While this is optional in practice, we find it helps in our experiments.

\textbf{Critic pessimism coefficient ($\rho$).}
We use a pessimistic critic backup \citep{dac_fang2025} in TD targets, computed from an ensemble of value functions, as in \citet{qam_li2026}. For an ensemble of $K$ value functions parameterized by $\phi_j$ for $j \in \{1, \dots, K\}$ and corresponding target networks parameterized by $\bar\phi_j$, the loss function is
\begin{align}
\gL(\phi_j) = \E_{\wt \tau}\big[\ell_2^\kappa\big(V_{\phi_j}(s, x^f, f) - (r + \gamma [ \bar{V}_{\text{mean}}(s', x'^0, 0) - \rho \bar{V}_{\text{std}}(s', x'^0, 0) ])\big)\big],
\end{align}
where $\bar{V}_{\text{mean}}(s', x'^0, 0) = \frac{1}{K} \sum_{k} V_{\bar{\phi}_k}(s', x'^0, 0)$, $ \bar{V}_{\text{std}}(s', x'^0, 0) = \sqrt{\frac{1}{K}\sum_{k} \left( V_{\bar{\phi}_k}(s', x'^0, 0) - \bar{V}_{\text{mean}}(s', x'^0, 0) \right)^2}$, and $\rho$ controls the degree of pessimism. See \Cref{sec:tuning} for more details.

\section{Experimental Details}
\label{sec:exp_details}

We evaluate all baselines with the official OGBench environments and datasets. We fix the following hyperparameters for fair comparison, unless otherwise mentioned.

\begin{table}[h!]
\caption{
\footnotesize
\textbf{Common Hyperparameters.}
}
\vspace{-5pt}
\label{table:gen_hypm}
\begin{center}
\scalebox{0.78}
{
\begin{tabular}{ll}
    \toprule
    \textbf{Hyperparameter} & \textbf{Value} \\
    \midrule
    Gradient steps & $2$M \\
    Optimizer & Adam~\citep{adam_kingma2015} \\
    Learning rate & $0.0003$ \\
    Batch size & $256$ \\
    MLP size & $[512, 512, 512, 512]$  \\
    Nonlinearity & GELU~\citep{gelu_hendrycks2016} \\
    Target network update rate & $0.005$ \\
    Flow steps $F$ & $10$ \\
    Discount factor $\gamma$ & $0.99$ (default), $0.995$ (\tt{humanoidmaze}, \tt{antmaze-giant}) \\
    Action chunking size $h$ & $1$ (locomotion), $5$ (manipulation) \\
    Ensemble size $K$ & $10$ \\
    Critic Target pessimistic coefficient $\rho$ & $0.5$ (default), $0$ (\tt{humanoidmaze}) \\
    \bottomrule
\end{tabular}
}
\end{center}
\end{table}

\subsection{Methods}
\label{sec:methods}

We implement RQL and provide commands to reproduce results at \url{https://github.com/aoberai/rql}.

\begin{itemize}

\item BDPO \citep{bdpo_gao2025}.
We use the original implementation by \citet{bdpo_gao2025} and sweep $\eta$ within $\{0.03, 0.1, 0.3, 0.7, 1\}$ and use $\rho=0.5$ for all tasks except \tt{humanoidmaze} which uses $\rho=0.0$. Following \citet{bdpo_gao2025}, BDPO requires an additional BC warmup stage, which we provide for $1$M additional offline steps. Thus, it uses twice as many offline steps as RQL and other methods in this comparison.

\item TFQL.
We implement TFQL and swept expectile $\kappa$  within $\{0.5, 0.7, 0.9\}$, and the BC regularization coefficient $\alpha$ from $\{0.1, 0.3, 1, 3, 10\}$. 

\item RQL.
See Appendix \ref{sec:tuning}.

\end{itemize}

For other methods, we use the official results by \citet{qam_li2026} and maintain an apples-to-apples setting when adding new results. Complete hyperparameters for BDPO, TFQL, and RQL can be found in Table \ref{table:hyp_task}.

\subsection{Hyperparameter Tuning}
\label{sec:tuning}

There are two important hyperparameters to tune when using RQL: expectile $\kappa$ and BC regularization $\alpha$. We swept expectile $\kappa$  within $\{0.5, 0.7, 0.9\}$, and the BC regularization coefficient $\alpha$ from $\{0.1, 0.3, 1, 3, 10\}$. While we use ensemble critic target pessimistic coefficient \citep{dac_fang2025} $\rho=0.5$ for all tasks except \tt{humanoidmaze} for apples-to-apples comparison with \citet{qam_li2026} results, we find $\rho=0$ is often better such as in \tt{cube-double}, and with additional tuning budget, we recommend ablating $\rho$ within $\{0, 0.5\}$. Similarly, we match \citet{qam_li2026} with action chunk horizon $h=5$ for manipulation and $h=1$ for locomotion tasks, but we recommend additional tuning within $\{1, 3, 5, 10\}$.

\begin{table}[h!]
\caption{
\footnotesize
\textbf{Task-specific hyperparameters.}
We describe task-specific hyperparameters below
($\alpha$: BC coefficient, $\kappa$: expectile, $\eta$: regularization strength, $\rho$: pessimistic coefficient).
}
\vspace{-5pt}
\label{table:hyp_task}
\begin{center}
\scalebox{0.78}
{
\begin{tabular}{lcccc}
\toprule
 & \tt{BDPO} & \tt{TFQL} & \tt{RQL} \\
\tt{Environment} & $(\eta, \rho)$ & $(\alpha, \kappa)$ & $(\alpha, \kappa)$ \\
\midrule
\tt{scene-sparse} & $(1, 0.5)$ & $(3, 0.7)$ & $(3, 0.7)$ \\
\tt{puzzle-3x3-sparse} & $(0.7, 0.5)$ & $(1, 0.5)$ & $(1, 0.7)$ \\
\tt{puzzle-4x4-sparse-100m} & $(0.7, 0.5)$ & $(3, 0.9)$ & $(1, 0.9)$ \\
\tt{cube-double} & $(0.7, 0.5)$ & $(10, 0.9)$ & $(10, 0.9)$ \\
\tt{cube-triple} & $(0.7, 0.5)$ & $(10, 0.9)$ & $(1, 0.9)$ \\
\tt{cube-quadruple-100m} & $(0.7, 0.5)$ & $(10, 0.9)$ & $(1, 0.7)$ \\
\tt{antmaze-large} & $(1, 0.5)$ & $(0.1, 0.7)$ & $(0.1, 0.5)$ \\
\tt{antmaze-giant} & $(1, 0.5)$ & $(0.1, 0.7)$ & $(0.1, 0.5)$ \\
\tt{humanoidmaze-medium} & $(0.3, 0)$ & $(0.3, 0.5)$ & $(0.3, 0.5)$ \\
\tt{humanoidmaze-large} & $(0.3, 0)$ & $(3, 0.7)$ & $(0.3, 0.5)$ \\
\midrule
\end{tabular}
}
\end{center}
\end{table}

\end{document}